\documentclass[letterpaper, 10 pt, conference]{ieeeconf}  % Comment this line out if you need a4paper

\IEEEoverridecommandlockouts                              % This command is only needed if 
\usepackage{graphicx} % for pdf, bitmapped graphics files
\usepackage{amsmath} % assumes amsmath package installed
\usepackage{amssymb}  % assumes amsmath package installed
\usepackage{subfigure}

\usepackage{xcolor}
\usepackage[acronym,toc]{glossaries}

\title{\LARGE \bf
Fine Tuning Swimming Locomotion Learned from Mosquito Larvae
}

\author{Pranav Rajbhandari$^{1}$ and 
Karthick Dhileep$^{2}$ and
Sridhar Ravi$^{3}$ and
Donald Sofge$^{4}$% <-this % stops a space
\thanks{$^{1}$Pranav Rajbhandari is the corresponding author and with Department
of Computer Science, Carnegie Mellon University, Pittsburgh, PA, USA, {\tt\small prajbhan@alumni.cmu.edu}. They completed this work under NREIP at Naval Research Laboratory, Washington D.C., USA. }
\thanks{$^{2,3}$Karthick Dhileep and Sridhar Ravi are with School of Engineering and Technology, University of New South Wales, Canberra, Australia.}
\thanks{$^{4}$Donald Sofge is with the Naval Research Laboratory, Washington D.C., USA, {\tt\small donald.a.sofge.civ@us.navy.mil}.}%
}

\usepackage{fancyhdr}
\begin{document}
\maketitle

\newcommand{\todo}[1]{\textcolor{red}{TODO: {#1}\\}}
\newcommand{\lrquote}[1]{\lq#1\rq}
\newcommand{\commentout}[1]{}

\def\fullfiguresize{.92\linewidth}
\def\halffiguresize{.49\linewidth}
\newacronym{RL}{RL}{Reinforcement Learning}
\newacronym{BGPS}{BGPS}{Baseline Guided Policy Search}
\newacronym{CFD}{CFD}{Computational Fluid Dynamics}
\newacronym{PPO}{PPO}{Proximal Policy Optimization}
\thispagestyle{fancy}
\pagestyle{fancy}

\begin{abstract}
In prior research, we analyzed the backwards swimming motion of mosquito larvae, parameterized it, and replicated it in a Computational Fluid Dynamics (CFD) model. 
Since the parameterized swimming motion is copied from observed larvae, it is not necessarily the most efficient locomotion for the model of the swimmer.
In this project, we further optimize this copied solution for the swimmer model. 
We utilize Reinforcement Learning to guide local parameter updates. 
Since the majority of the computation cost arises from the CFD model, we additionally train a deep learning model to replicate the forces acting on the swimmer model. 
%We use this CFD clone to perform Model Based RL.
We find that this method is effective at performing local search to improve the parameterized swimming locomotion.
\end{abstract}
\section{Introduction/Related Work}\subsection{Locomotion of Mosquito Larvae}
In previous research, we parameterize the swimming motion of mosquito larvae and successfully replicate it inside a computational fluid dynamics simulator \cite{mosq}.
We model the swimmer as a 2D boundary and use the immersed boundary lattice Boltzmann method (IB-LBM) \cite{flow_calc} to calculate forces and resulting trajectory of a swimming locomotion.

\begin{figure}[htbp!]
    \centering
\includegraphics[width=\fullfiguresize]{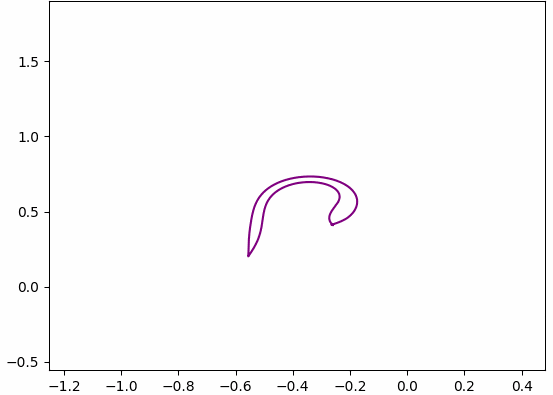}
\caption{Model of 2D swimmer in CFD}
\label{cfd_swimmer}
\end{figure}

For the parametrization, we discretize the swimmer into line segments and estimate the angle $\theta$ between adjacent segments.
This angle varies with time $t\in\mathbb R$ as well as location on the swimmer $s\in\mathbb R$. 
We found that $\theta(s,t)$ was well approximated by Equation \ref{parametrized_swimming}, using an amplitude function $\theta_0(s)$, a frequency $\omega$, and phase shift function $\phi(s)$. 
\begin{equation}
    \label{parametrized_swimming}
    \theta(s,t)=\theta_0(s)\cdot\sin(\omega t+\phi(s))
\end{equation}
We approximate $\theta_0$ and $\phi$ as polynomials with respect to $s$ of degrees 5 and 4 respectively.
In addition to $\omega$, this results in a 12 dimensional parameter space.

Explicitly, we may rewrite Equation \ref{parametrized_swimming} using a parameter vector $\mathbf p\in\mathbb R^{12}$:
\begin{equation}
    \label{parameter_stuff}
    \theta_{\mathbf{p}}(s,t)=
    \left(\sum\limits_{i=0}^5 s^i\mathbf{p}_{i+1}\right)\cdot\sin\left(\mathbf{p}_{12}t +\sum\limits_{i=0}^4 s^i\mathbf{p}_{i+7}\right)
\end{equation}

We obtain our initial parameters in \cite{mosq} by estimating the motion of live mosquito larvae.
\subsection{Local Search/Hill Climbing}
The hill climbing algorithm is a well-known local search method that repeatedly updates a solution to an improvement found by testing a local neighborhood \cite{aimodern}. 
In continuous search spaces, this can be approximated by fixing a step size $\delta$ and searching around a solution by taking a $\delta$ step in every dimension. 
This approximates a gradient of the objective with respect to the parameter space, and this approximation can be done in $O(d)$ for $d$ the number of dimensions. 

We may apply this to optimizing the parameters of an initial swimming locomotion. 
We set our objective to displacement in some set time, and evaluate a solution through a simulation.
With this method, a single update (assuming we take a full gradient estimation) will require $O(d)$ simulations. 

This is a reasonable approach if we utilize our simulation only for evaluating a potential solution. 
However, by making small adjustments to the swimming policy mid-episode, we can better estimate which updates increase our objective.
To make these adjustments, we utilize a \gls{RL} algorithm.

\subsection{Baseline Guided Policy Search}
Hu and Dear explore a similar problem of training an articulated robotic swimmer through \gls{RL} \cite{swimmy}.
They introduce \gls{BGPS}, an augmented \gls{RL} algorithm which starts at an approximated policy and allows an agent to add small adjustments. 
In their research, they utilize this method to optimize swimming motion in robotic swimmers composed of three segments. 

We utilize this technique to make adjustments mid-simulation to a swimming locomotion. 
We will then learn parameters that best approximate this adjusted policy, updating the baseline.

Since \gls{BGPS} is restricted to making relatively small updates to the swimming motion, we expect that when projected to parameter space, the update will be relatively small. 
Thus, in parameter space, this will behave similarly to a local search algorithm. 
The main distinction is the number of samples an update requires. 
We hypothesize that a \gls{RL} algorithm will learn kinematic information to find an improving update within a simulation or two. 
In contrast, a standard local search would need to sample simulations of more neighboring solutions to make an update. 

%\section{Related Work}\input{parts/relate}
\section{Methods}\subsection{Simulated Mosquito Swimmer}
In previous research, we create a simulated mosquito larvae inside a \gls{CFD} simulator. The setup is able to replicate the dynamics of real mosquito larvae given the correct swimming motion. We utilize this \gls{CFD} model to fine tune the locomotion of the same simulated swimmer.

\subsection{RL-Guided Parameter Update}

\begin{figure}[ht!]
    \centering
\includegraphics[width=\fullfiguresize]{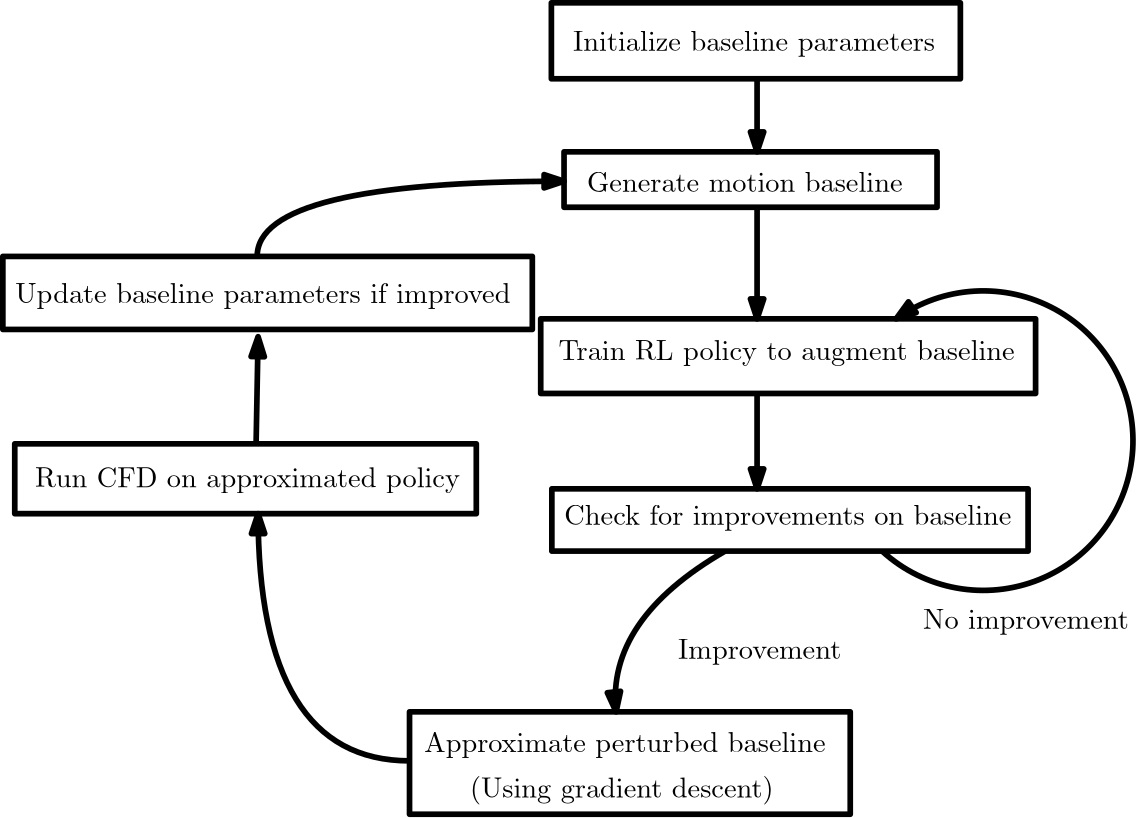}
\caption{Local parameter search using BGPS algorithm}
\label{bgps}
\end{figure}

We implement a RL environment utilizing a \gls{CFD} simulation. 

We use this environment to optimize a set of parameters as in Figure \ref{bgps}. 
We first repeatedly run the BGPS algorithm, searching for an augmentation that outperforms the baseline policy.
Once this policy is found, we approximate parameters that match the augmented policy. 
Finally, if these updated parameters are truly an improvement, we update the baseline and continue with our loop.

To approximate parameters, we inspect the augmented policy and record the angles $\theta^*(s,t)$ between adjacent line segments. 
We sample values of $s$ that constitute each joint of the swimmer, and values of $t$ within one period of the swimming motion. 
We then must choose parameters $\mathbf{p}$ such that $\theta_{\mathbf p}(s,t)$ from Equation \ref{parameter_stuff} approximates $\theta^*(s,t)$.
Since $\theta_{\mathbf p}$ is differentiable with respect to $\mathbf p$, we do this through gradient descent, minimizing the Mean Squared Error loss (Equation \ref{mseloss}).
We initialize our search with the baseline parameters, since $\theta^*(s,t)$ results from small adjustments to this. 

\begin{equation}
    \label{mseloss}
    \mathcal L(\mathbf p)=\mathbb{E}_{s,t}\left[\left(\theta^*(s,t)-\theta_{\mathbf p}(s,t)\right)^2\right]
\end{equation}

\subsection{CFD Clone}
We utilize deep learning to create a model that predicts the forces acting on a simulated swimmer based on its outline. We experiment with both sequential models and a normal feed forward network.

We use \gls{CFD} on a parameter sweep of parameterized swimming motions to create the training data. To define the model loss, we use the sum of mean squared error loss and cosine similarity loss to further ensure the forces are in the correct direction.

\subsubsection{Network Input}
We allow the network to observe the COM-centered outline of the swimmer at each timestep. This is a set of 400 sampled points on the swimmer surface. We use this as our network input since it is the same input to the \gls{CFD} model. 
For our feed-forward network, we additionally allow the network to observe the past 3 timesteps to get kinematic information about the swimmer. 
\subsubsection{Network Output}
The network output is the surface forces on each of the 400 sampled points. We use this as our network output since it is output of the \gls{CFD}, and it is sufficient to calculate the movement of the swimmer.
\subsubsection{CFD calculation}
We use the trained model to create a \gls{CFD} clone by calculating surface forces at every timestep and applying kinematic equations, similar to the calculations in Zhu et al. \cite{kine}.

\section{Experiments}\subsection{CFD Clone}
We experimented with the seq-to-seq Recursive Neural Network (RNN) model \cite{rnn} and the Long Short-Term Memory (LSTM) model \cite{lstm}. 
We also included a residual network for comparison with non-sequential methods. 

We hypothesize that sequential models are better suited to handle the estimation of forces on our swimmer. 
Our reasoning is that a non-sequential approach would suffer from noise in the training data, as an estimate must be made from information in just a few time steps. 
In contrast, a sequential model can obtain information from the full history of the swimmer, allowing it to be more robust to the noise.

\subsubsection{Network Architecture}
In addition to varying the model used, we also evaluate different network sizes in their ability to reduce the objective function. We take our best performing model and vary the depth of the architecture from one layer to eight layers. In our final CFD clone, we use the simplest network that performs comparably well.

\subsection{Baseline Guided Policy Search}

We implement BGPS to make adjustments to a baseline swimming policy. 
As in Figure \ref{bgps}, we alternate between using BGPS to improve the baseline and fitting parameters to the augmented policy.
We use the stable\_baselines3 \cite{stable-baselines3} implementation of \gls{PPO}, a standard on-policy \gls{RL} algorithm \cite{ppo}.

\subsubsection{Observation Space}
Guided by Hu and Dear's work \cite{swimmy}, we allow the agent to observe the current time (encoded periodically by applying sine and cosine at various frequencies), the angles of a few points on its midline, its heading angle, and its position and velocity.
\subsubsection{Action Space}
The action space available to the agent is a list of angles corresponding to joints on its midline. 
The angles output from the RL agent are added onto a baseline policy.
\subsubsection{Rewards}
We observe the normalized total displacement of the baseline policy's movement.
At each timestep, we give the \gls{RL} agent rewards equivalent to the displacement in the direction of the baseline displacement vector. 
We do this to ensure that the sum of rewards in an episode is the swimmer's overall displacement in the same direction as the baseline policy. 

\section{Results}\subsection{CFD Clone}

\begin{figure}[ht!]
    \centering

\includegraphics[width=\fullfiguresize]{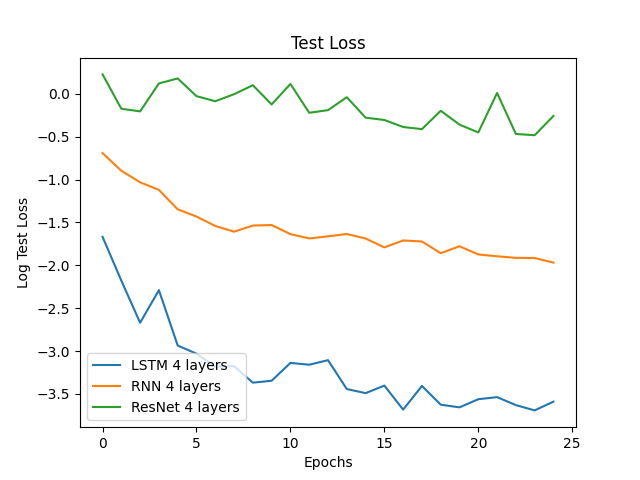}
\caption{Comparison of log test losses of various types of models}
\label{4plot}
\end{figure}

We inspect the test loss of the LSTM, RNN, and residual network models during training. 
We find that LSTM performed the best, RNN was second best, and the Residual network performed the worst (Figure \ref{4plot}).
The fact that both sequential models outperformed the residual network indicates that these models are better suited to estimate kinematic information throughout an episode. 

Since the LSTM model performed the best, we proceed to evaluate the performance of various LSTM network sizes. 

\subsubsection{Network Architecture}

\begin{figure}[ht!]
\centering
\includegraphics[width=\fullfiguresize]{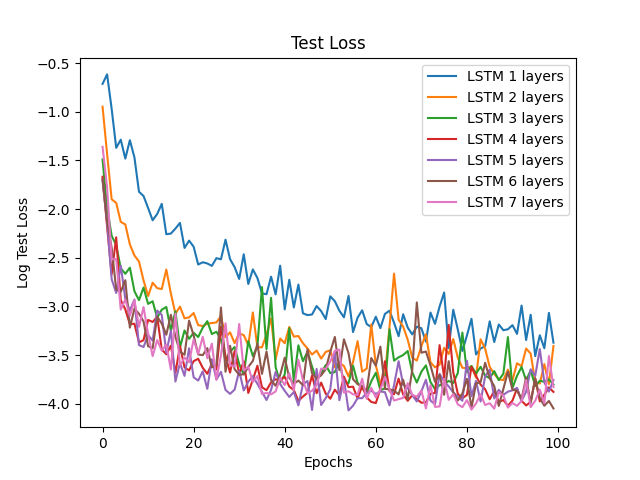}
\caption{Comparison of log test losses of various LSTM network depths}
\label{lstmplot}
\end{figure}
We test and compare LSTM networks ranging from one to seven layers in depth. 
We notice that at depths of 1 and 2 the networks perform worse with respect to their test loss  (Figure \ref{lstmplot}). 
The networks at higher depths all perform similarly.
Since a depth of 3 is the shallowest network that well compared to all other network depths, we use this depth in our CFD clone. 
\subsection{BGPS}
\begin{figure}[ht!]
    \centering
\includegraphics[width=\fullfiguresize]{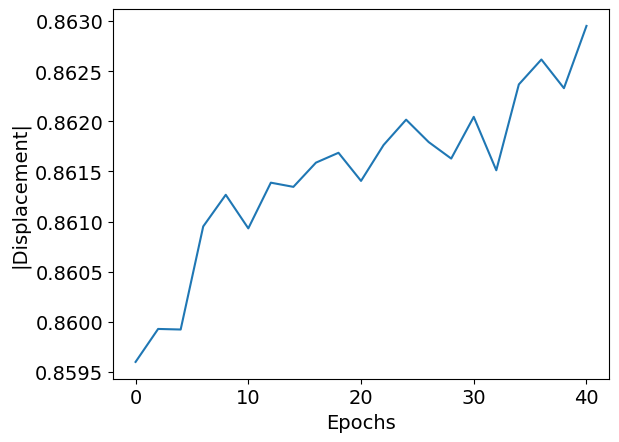}
\caption{Result of BGPS on swimmer displacement per episode}
\label{resultplot}
\end{figure}

We use the resulting CFD clone to optimize our swimming locomotion from the initial choice of parameters. 
In each episode of training, we record the absolute value of the total displacement. 

From Figure \ref{resultplot}, we find that the BGPS algorithm is successful in gradually optimizing the movement of the simulated swimmer.

However, the scale of the improvement is small in comparison to the size of the displacement. 
This could be a result of the scale we allow BGPS to augment the policy. 

\section{Conclusion}In this study, we fine tune a learned parameterized swimming locomotion for a specific platform.
We use a local search to gradually update the parameters towards more optimal neighbors. 
To increase efficiency, we use \gls{RL} to learn kinematic information about the swimming locomotion, guiding the local search.

We additionally approximate the learning environment with a \gls{CFD} clone, learned through a deep neural network. 
We utilize this \gls{CFD} clone to efficiently conduct model-based RL to improve the baseline policy. 
Overall, we take advantage of kinematic nature of our optimization problem to improve the speed of local search. 

We find that these methods are successful in improving the parameterized swimming locomotion through local search.
However, we find that the scale of the improvements are small. 

In future research, we plan to vary the amount that BGPS can augment the policy to obtain more drastic differences. 
We also plan to use this method to optimize locomotion on a physical robotic swimmer.

\bibliographystyle{plain} 
\bibliography{refs}

\begin{thebibliography}{1}

\bibitem{mosq}
Karthick Dhileep, Qiuxiang Huang, Fangbao Tian, John Young, Joseph~C.S. Lai, Donald Sofge, and Sridhar Ravi.
\newblock {Investigation of bio-inspired tail-first swimming using numerical and robotic models}.
\newblock In {\em 2023 IEEE International Conference on Robotics and Biomimetics (ROBIO)}, pages 1--6, 2023.

\bibitem{lstm}
Sepp Hochreiter and Jürgen Schmidhuber.
\newblock Long short-term memory.
\newblock {\em Neural Computation}, 9(8):1735--1780, 1997.

\bibitem{swimmy}
Jiaheng Hu and Tony Dear.
\newblock Guided deep reinforcement learning for articulated swimming robots, 2023.

\bibitem{flow_calc}
Qiuxiang Huang, Zhengliang Liu, Li~Wang, Sridhar Ravi, John Young, Joseph Lai, and Fang-Bao Tian.
\newblock {Streamline penetration, velocity error, and consequences of the feedback immersed boundary method}.
\newblock {\em Physics of Fluids}, 34(9), 2022.

\bibitem{stable-baselines3}
Antonin Raffin et~al.
\newblock {Stable-Baselines3: Reliable Reinforcement Learning Implementations}.
\newblock {\em Journal of Machine Learning Research}, 22(268):1--8, 2021.

\bibitem{rnn}
David~E. Rumelhart and James~L. McClelland.
\newblock {\em Learning Internal Representations by Error Propagation}.
\newblock MIT Press, 1987.

\bibitem{aimodern}
Stuart Russell and Peter Norvig.
\newblock {\em {Artificial intelligence: a modern approach}}.
\newblock Pearson, 2016.

\bibitem{ppo}
John Schulman et~al.
\newblock {Proximal Policy Optimization Algorithms}, 2017.

\bibitem{kine}
Yi~Zhu, Fang-Bao Tian, John Young, James Liao, and Joseph Lai.
\newblock A numerical study of fish adaption behaviors in complex environments with a deep reinforcement learning and immersed boundary–lattice boltzmann method.
\newblock {\em Scientific Reports}, 11:1691, 01 2021.

\end{thebibliography}

\end{document}